\documentclass[10pt,conference,a4paper]{IEEEtran}
\usepackage{graphicx}
\graphicspath{{figures/}}
\usepackage{amsmath}
\usepackage{amssymb}
\usepackage{booktabs}
\usepackage{multirow}
\usepackage[table,xcdraw]{xcolor}
\usepackage{hyperref}

\ifCLASSINFOpdf
\else
\fi
\begin{document}
	%
	\title{A Dual-branch Network for \\Infrared and Visible Image Fusion}

	\author{\IEEEauthorblockN{Yu~Fu}
		\IEEEauthorblockA{Jiangsu Provincial Engineering\\
			Laboratory of Pattern Recognition and\\
			Computational Intelligence,\\
			Jiangnan University,\\
			Wuxi, China, 214122\\
			Email:  yu\underline{~}fu\underline{~}stu@outlook.com}
		\and
		\IEEEauthorblockN{Xiao-Jun Wu}
		\IEEEauthorblockA{Jiangsu Provincial Engineering\\
			Laboratory of Pattern Recognition and\\
			Computational Intelligence,\\
			Jiangnan University,\\
			Wuxi, China, 214122\\
			Email:xiaojun\underline{~}wu\underline{~}jnu@163.com}
		\and
		}
	
	
	%


	\maketitle
	
	\begin{abstract}
	In recent years, deep learning has been used extensively in the field of image fusion. In this article, we propose a new image fusion method by designing a new structure and a new loss function for a deep learning model. Our backbone network is an autoencoder, in which the encoder has a dual branch structure. We input infrared images and visible light images to the encoder to extract detailed information and semantic information respectively. The fusion layer fuses two sets of features to get fused features. The decoder reconstructs the fusion features to obtain the fused image. We design a new loss function to reconstruct the image effectively. Experiments show that our proposed method achieves state-of-the-art performance.

	\end{abstract}
	

	%
	\IEEEpeerreviewmaketitle

	\section{Introduction}
	Infrared and visible light image fusion is an important subject in the field of image processing. Infrared images and visible light images come from different signal sources. 
	Generally, visible light images have high spatial resolution and rich detail information. Meanwhile, infrared images contain rich thermal radiation information and global structure information.
	How to extract important information of source images from different sources is a problem that needs to be solved. A high-quality fused image can be used for video surveillance, object recognition, tracking and remote sensing, etc.
	
	In the field of traditional methods, image fusion has achieved amazing results. Some classic or popular image fusion algorithms are introduced by Li et al. \cite{li2017pixel}. Here are some mature multi-scale transform fusion methods. Source images are decomposed into a set of multi-scale representation features and fused by pyramid\cite{mertens2009exposure}, curvelet\cite{zhang1999categorization}, contourlet\cite{upla2014edge} , etc. Inverse multi-scale transform is used to obtain the fused image.
	In the low-rank representation domain, Li and Wu et al. proposed MDLatLRR\cite{li2020mdlatlrr} based on deep decomposition with latent LRR which extracts the features of image in the low-rank domain. In the Subspace domain, there are successful algorithms such as PCA\cite{pca2017}, ICA\cite{ICA2007}, NMF\cite{mou2013image}.
	
	With the development of deep learning in recent years, the powerful feature extraction capabilities of deep neural networks have become increasingly prominent. Liu et al. summarized many outstanding CNN-based image fusion methods\cite{liu2018deep}. 
	For example, features are extracted using dense block (DenseFuse) proposed by Li and Wu \cite{li2018densefuse} or PCA filters of PCANet\cite{song2018multi} proposed by Song and Wu. The extracted features are fused with specific fusion strategies.
	In addition to using neural networks as feature extraction tools, there are also end-to-end fused image generation networks such as FusionGan\cite{ma2019fusiongan} proposed by Ma et al.
	
	Compared with the end-to-end network, we think that using deep neural networks to extract features reasonably can better perform image fusion tasks. Moreover, 
	deep neural networks have more powerful feature extraction capabilities than traditional methods. However, the existing fusion networks have only several layers. We try to design a more reasonable network structure to extract their important features from infrared and visible light images.

	In this paper, we propose a novel and effective autoencoder in which encoder has dual branch. One branch is the detail branch, which uses dense connections for extracting shallow and edge information. Another branch is the semantic one, which uses rapid downsampling  extracting semantic and structural information. 
	The infrared and visible light images are input to the encoder to obtain two sets of features containing the original image information. Two sets of features are fused by fusion layer. The decoder reconstructs the fused feature to obtain the fused image.
	
	This paper is structured as follows. In Section II, we introduce the related image fusion algorithms. Section III presents the network structure, loss function, and the fusion strategy respectively. Section IV introduces the training details and the results of its comparison with state-of-the-art methods. Section V draws the paper to conclusion.
	\begin{figure*}{}
		\includegraphics[width=\textwidth]{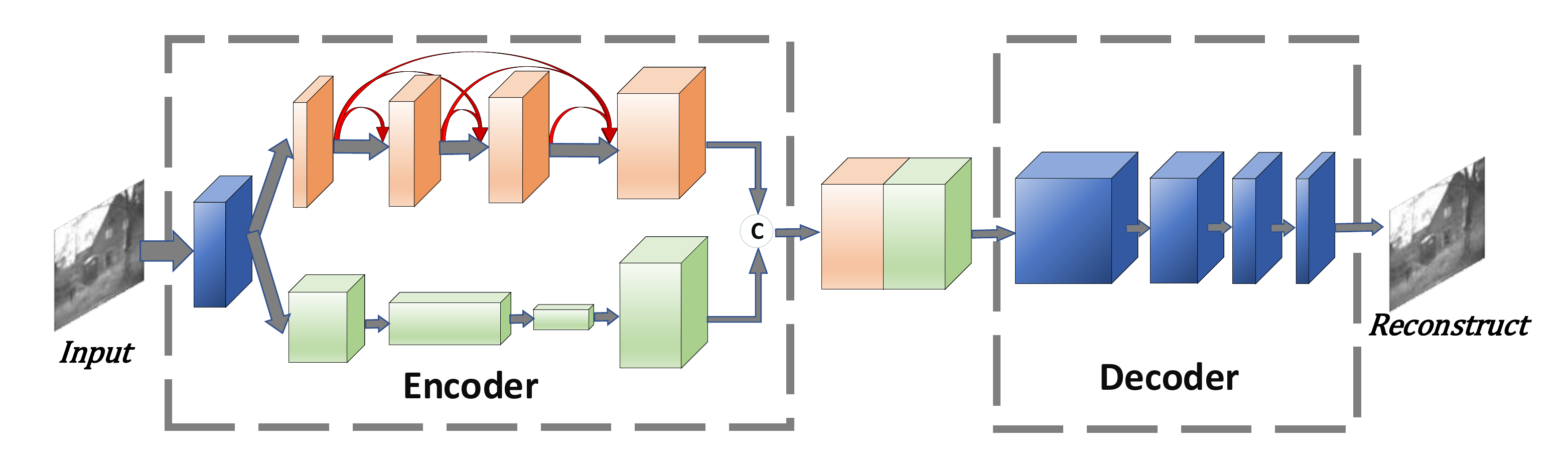}
		\caption{The backbone of our network in this paper. The orange blocks is the detail branch, and the green blocks is the semantic branch. Concatenate the encoding results of the two branches to get our latent layer.
		} \label{fig:backbone}
	\end{figure*}
	\section{Related Work}
	In recent years, the development of deep learning has promoted the new methods of image fusion.
	
	In 2017, Liu proposed a multi-focus image fusion based on CNN\cite{liu2017multi-focus}, which divides the image into many small blocks, and the network is utilized to predict whether the block is clear or fuzzy, and multi-focus fusion based on the predicted decision map is developed. This method simply uses the classification capabilities of CNN,  and firstly using CNN in the field of image fusion, but cannot be used in the field of infrared images and visible light images.
	
	Ma et al. proposed an end-to-end image generation network FusionGan\cite{ma2019fusiongan} in 2019. They use the generator to generate the fused image, and use the discriminator for adversarial learning. However, the image generated by GAN is unstable and has a lot of noise sometimes.
	
	In ICCV 2017, Prabhakar et al. proposed Deepfuse\cite{prabhakar2017deepfuse} for multi-exposure image fusion. They proposed a simple CNN structured autoencoder with a total of five layers. They use the encoder to extract features and use the decoder to reconstruct the fused image. 	
	Huang and Liu et al. propose the DenseNet\cite{huang2017densely}. In the network, the input of each layer are concatenated with the outputs of previous layers. Denseblock strengthens the forward propagation of features and the reuse of features, so that deep networks can also obtain shallow information.	
	Li and Wu  proposed Densefuse\cite{li2018densefuse} for infrared and visible light image fusion in 2019. Compared with the network structure of Deepfuse, they increased the number of network layers and added denseblock structure and designed a new fusion strategy.

	\section{The Proposed Fusion Method}
	In this section, we introduce the designed deep neural network in detail. This section gives details of the network, loss function, and fusion strategy.

	\subsection{Network Structure}
	We design an autoencoder network to reconstruct the image. This network includes two parts: encoder, and decoder. The proposed network structure is shown in Fig.\ref{fig:backbone}.

	In the training phase, we input infrared and visible light images $I(I_1,I_2,I_3...I_k)$ into the network. The input images will be resized to a fixed size. Then, the images are processed by the encoder and decoder. The different types of images share one same encoder and decoder.
	
	Our encoder has two branches, the detail branch and the semantic branch. In the encoder, we first perform convolution on the input images $I$ to obtain a set of features. Then this set of features goes through the detail branch and the semantic branch at the same time. 
	
	The purpose of the detail branch of the encoder is to extract more detailed and texture space information from the original image. Therefore, we design the detail branch as a four-layer denseblock structure. As shown in the upper branch of the encoder in the Fig.\ref{fig:backbone}. The input of each layer of convolution concatenates with the outputs of all previous layers. That is, the output of $p_{th}$ layer is $X_p= F_p ([X_0, X_1, X_2,... , X_{p-1}])$ . Where $F_p$ represents a nonlinear operation of the $p_{th}$ layer including normalization, activation layer, convolution layer, etc.  $[X_0, X_1, X_2,... , X_{p-1}]$ represents concatenation of the outputs of all prior layers. Such a densely connected network can enable the detail branch to better learn the shallow features of the original images. This branch can focus on low-level details.
	
	In addition to detail information, of course we also need global structural semantic information. In order to extract the global information of the features, we design a rapid downsampling network structure, also called semantic branch. As shown in the lower branch of the encoder in the Fig. \ref{fig:backbone}. We quickly downsample the features three times and expand the number of channels of the features. Then the semantic features is upsampled to the original size to be able to concatenate with the detail features.
	\begin{figure}
		\includegraphics[width=\linewidth]{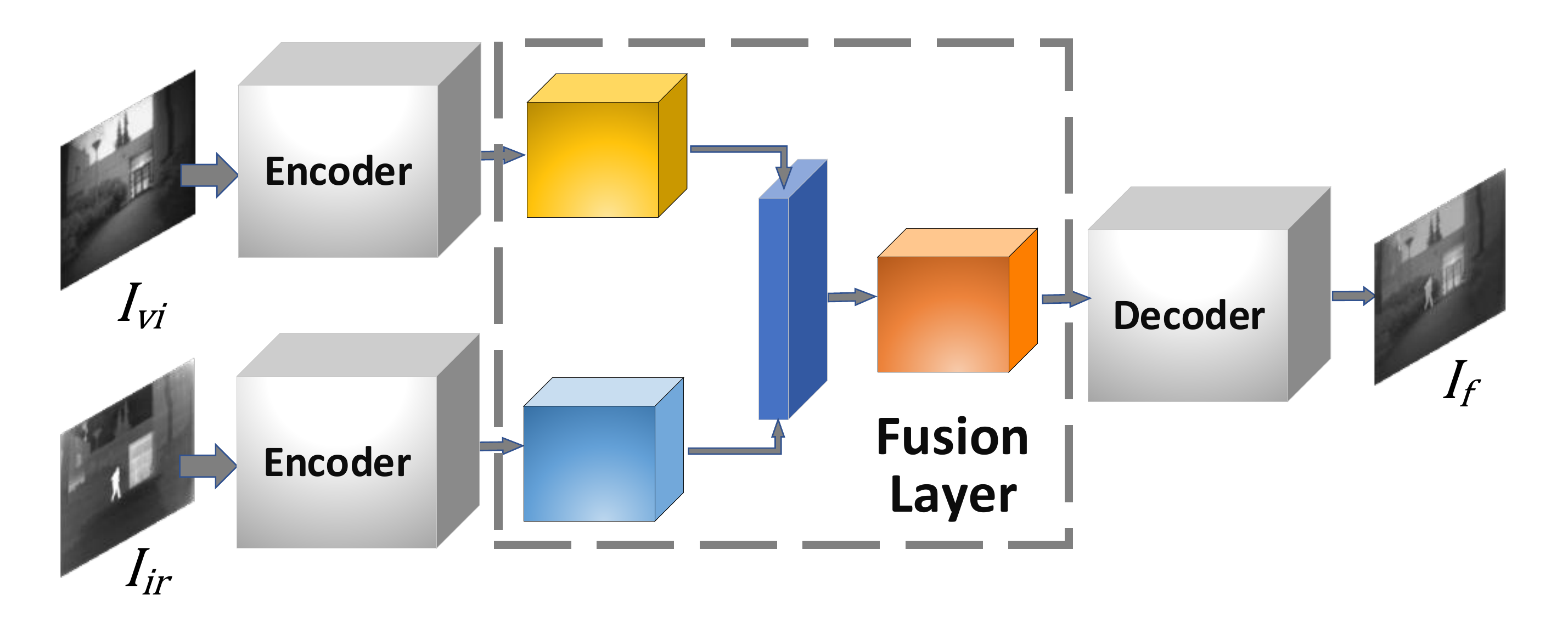}
		\caption{The network during the test phase.
		} \label{fig:test}
	\end{figure}
	During the test phase, we input a pair of the infrared and visible light images into the weights shared siamese encoder at the same time. Then, we can obtain two set of features separately. We design a fusion layer in the test phase to fuse these features. Finally, we input the fused features into the decoder to reconstruct the fused image, as shown in the Fig. \ref{fig:test}. The details about the fusion layer will be shown in the later sections.
	
	\subsection{The details of the network}
	The input images will be pre-registered to resize to 128$\times$128. Firstly, we perform a convolution on the images and obtain 32 channels of features. Then we make a copy of these features and input them into the detail branch and semantic branch respectively. 
	
	In the detail branch, we perform four convolution and dense connection operations. The size of features is unchanged, and the number of channels is accumulated, which is 16, 32, 48 and 64. In the semantic branch, we perform four convolution operations with a step of 2 to achieve the propose of downsampling. The size of the features is reduced to 64, 32 and 16. The number of channels is first increased to 64 and 128, and then decreased to 64. Therefore, the number of channels can be the same as the detail branch. At the end of the semantic branch, we upsample the features using the bilinear method, expanding their size by 8 times.
	
	Finally, after four convolutions, the number of channels is gradually reduced to one in the decoder. And we can obtain the reconstituted images or the fused images. 
	All the above convolution operations use 3$\times$3 kernels with padding. The parameters of the network are shown in Table. \ref{parameters}.
	
	\begin{table}[]
		\caption{The parameters of the network.  }\label{parameters}
		\begin{center}
			\resizebox{\linewidth}{!}{
			\begin{tabular}{@{}|c|cccccc|@{}}
				\toprule
				\multicolumn{2}{|c|}{\multirow{2}{*}{Block}}                  & \multirow{2}{*}{Layer} & Channel & Channel  & Size    & Size     \\
				\multicolumn{2}{|c|}{}                                        &                        & (input) & (output) & (input) & (output) \\ \midrule
				\multirow{9}{*}{Encoder} &                                    & Conv\_1                & 1       & 32       & 128$\times$128 & 128$\times$128  \\ \cmidrule(l){2-7} 
				& \multirow{4}{*}{\begin{tabular}[c]{@{}c@{}}Detail\\ Branch\end{tabular}}   & Conv\_d1               & 32      & 16       & 128$\times$128 & 128$\times$128  \\
				&                                    & Conv\_d2               & 16      & 16       & 128$\times$128 & 128$\times$128  \\
				&                                    & Conv\_d3               & 32      & 16       & 128$\times$128 & 128$\times$128  \\
				&                                    & Conv\_d4               & 48      & 16       & 128$\times$128 & 128$\times$128  \\ \cmidrule(l){2-7} 
				& \multirow{4}{*}{\begin{tabular}[c]{@{}c@{}}Semantic\\ Branch\end{tabular}} & Conv\_s1               & 32      & 64       & 128$\times$128 & 64$\times$64    \\
				&                                    & Conv\_s2               & 64      & 128      & 64$\times$64   & 32$\times$32    \\
				&                                    & Conv\_s3               & 128     & 64       & 32$\times$32   & 16$\times$16    \\
				&                                    & Upsample               & 64      & 64       & 16$\times$16   & 128$\times$128  \\ \midrule
				\multicolumn{2}{|c|}{\multirow{4}{*}{Decoder}}                & Conv\_1                & 128     & 64       & 128$\times$128 & 128$\times$128  \\
				\multicolumn{2}{|c|}{}                                        & Conv\_2                & 64      & 32       & 128$\times$128 & 128$\times$128  \\
				\multicolumn{2}{|c|}{}                                        & Conv\_3                & 32      & 16       & 128$\times$128 & 128$\times$128  \\
				\multicolumn{2}{|c|}{}                                        & Conv\_4                & 16      & 1        & 128$\times$128 & 128$\times$128  \\ \bottomrule
			\end{tabular}
		}
		\end{center}
	\end{table}
	
	The ReLU function discards negative activation and loses half of the activation information. It may be that the sparse ability of the ReLU function is suitable for image classification, but not suitable for image reconstruction. LeakyReLU solves this problem to a certain extent, and can retain certain negative activations. We choose the MISH  function\cite{misra2019mish} as the activation function, which is more sensitive to small errors and has a suppressive effect on the large errors. The formula is as follows:
	\begin{equation}
	Mish(x) = x\cdot tanh(ln(1+e^x))
	\end{equation}

	\subsection{Loss Function}
	In order to better extract detailed information and semantic information, we design multiple losses in $L$, including pixel loss, gradient loss, color loss and perceptual loss respectively. The formula is presented as follows:
	\begin{equation}
	\label{equ:loss}
	L=L_{pixel}+\alpha L_{gradient}+\beta L_{color} +\gamma L_{perceptual}
	\end{equation}
	where $\alpha$, $\beta$, $\gamma$ are hyper parameters, used to balance the weight of the four losses.
	
	$L_{pixel}$ loss function can calculate the pixel error of the reconstructed image and the input image, the formula is given as follows:
	\begin{equation}
	L_{pixel} = MSE(I_{re},I_{in})
	\end{equation}
	\begin{equation}
	MSE(X,Y)=\frac{1}{N}\sum_{n=1}^{N}(X_n - Y_n )^2 
	\end{equation}
	where $I_{re}$ is the reconstructed image, $I_{in}$ is the input image, and $MSE(x,y)$ is the mean square error function of $x$ and $y$. 
	
	$L_{gradient}$ loss function can calculate the edge information loss of the reconstructed image and the input image, the formula is presented as follows:
	\begin{equation}
	L_{gradient} = MSE(Gradient(I_{re}),Gradient(I_{in}))
	\end{equation}
	where, $Gradient(x)$ is the image sharpening using the Laplace operator to obtain the gradient map. The Laplace operator performs a mathematical convolution operation. Its definition and approximate discrete expression is given as follows:
	\begin{equation}
	\begin{small}
	\begin{aligned}
	&\nabla^2f(x,y)=\frac{\partial^2 f(x,y)}{\partial x^2}+
	\frac{\partial^2 f(x,y)}{\partial y^2}\\
	&\approx [f(x+1,y) + f(x-1,y)+f(x,y+1)f(x,y-1)]-4f(x,y)
	\end{aligned}
	\end{small}
	\end{equation}
	
	$L_{color}$ loss function function can calculate the color histogram error of the reconstructed image and the input image, because we consider the light and dark of the visible light image and the infrared radiation information of the infrared image.
	\begin{equation}
	L_{color} =\frac{1}{255}\| Histogram(I_{re})-Histogram(I_{in})\|_2
	\end{equation}
	where $Histogram(x)$ is the color histogram of $x$. We set the number of the histogram bins to 255, the maximum and minimum values of the histogram calculation are the maximum and minimum values between $I_{re}$ and $I_{in}$. We calculate the norm of the difference between the two histograms.
	
	$L_{perceptual}$ loss function\cite{johnson2016perceptual} can calculate the errors of features between the reconstructed image and the input image.
	\begin{equation}
	L_{perceptual} = \sum_{i=1}^{4}MSE(\phi_i(I_{re}),\phi_i(I_{in}))
	\end{equation}
	where $\phi_i(x)$ represents the features of the $i$th layer obtained by inputting images $I$ into a specific trained network. In this paper, we choose pre-trained vgg19 network as the feature extraction network.
	\begin{equation}
	\end{equation}
	
	\begin{figure*}
		\includegraphics[width=\textwidth]{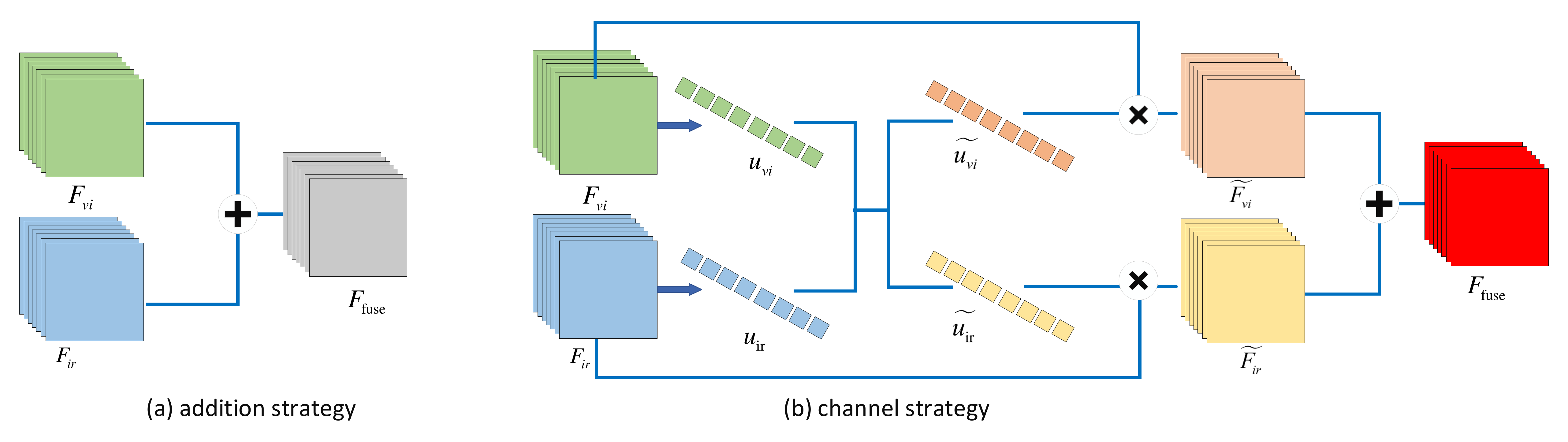}
		\caption{The network during the test phase. 
		} \label{fig:strategy}
	\end{figure*}

	\subsection{Fusion Strategy}
	We choose two feature fusion strategies, one is addition strategy and the other is channel strategy, as shown in Fig. \ref{fig:strategy}:
	
	\subsubsection{Addition Strategy}
	This is a simple and effective fusion method. Two sets of features are element-wisely added in the channel dimension as follows.
	\begin{equation}
	F_{fuse}(x,y) = F_{ir}(x,y)+F_{vi}(x,y)
	\end{equation}
	where $F_{fuse}$, $F_{ir}$ and $F_{vi}$ represent the fused features, infrared features and visible light features, respectively. $(x, y)$ denotes the corresponding position of features.
	
	\subsubsection{Channel Strategy}
	It is easy to know that the features of infrared images and visible light images have different importance on different channel. The visible light images contain more detail information, while the infrared images contain more semantic structural information. Moreover, the extracted features of different channels generated by different kernels are also different. 
	
	We design a channel selection strategy.
	As shown in (b) of Fig. \ref{fig:strategy}, the infrared and visible image features $F_{ir},F_{vi},F \in \mathbb{R}^{H \times W \times C}$ are pooled by the global average to $u_{ir},u_{vi},u \in \mathbb{R}^{1 \times 1 \times C}$. We calculate the probability $\widetilde u_{ir},\widetilde u_{vi}$ based on $u_{ir},u_{vi}$. Multiplying  $F_{ir}$ and $\widetilde u_{ir}$, $F_{vi}$ and $\widetilde u_{vi}$ to get the enhanced channel features $\widetilde F_{ir},\widetilde F_{vi}$ respectively. Element-wise adding $\widetilde F_{ir}$ and $\widetilde F_{vi}$ to get the final fusion features ${ {F}}_{fuse}$.The formula are given as follows:
	\begin{equation}
	\begin{aligned}
	& {u}_i=\frac{1}{H \times W}\sum_{m=1}^{H} \sum_{n=1}^{W} x_{i}(m, n) \\ 
	&\widetilde{ {u}}_{ir}=\frac{e^{ u_{ir}}}{e^{u_{ir}}+e^{u_{vi}}}~,~
	\widetilde{ {u}}_{vi}=1-\widetilde{ {u}}_{ir} \\
	&\widetilde{ {F}}_{ir} = \widetilde{ {u}}_{ir} {F_{ir}}~,~~~~
	\widetilde{ {F}}_{vi} = \widetilde{ {u}}_{vi} {F_{vi}}\\
	& {F}_{fuse}  = \widetilde{ {F}}_{ir} + \widetilde{ {F}}_{vi}
	\end{aligned} 
	\end{equation}
	\section{Experiments and Analysis}
	
	\subsection{Training and Testing Details}
	For the selection of hyperparameters, we make the values of four losses as close to the same order of magnitude as possible. So, in formula \ref{equ:loss}, we set $\alpha=1, \beta=0.001, \gamma=1000 $ by cross validation..
	
	We select 59 pairs of images in the TNO\cite{toet2014tno} data set as training and test data sets. We select 10 pairs of them as the test set and 49 pairs of them as the training set. 49 pairs of images are not enough as a training set, we crop each image to a size of 128$\times$128 with a step size of 14. At the same time, the cropped images will be expanded symmetrically. Finally we get 6,720 images as training set. 
	
	In the training phase, the batchsize = 64 images is set for training. We choose Adam \cite{kingma2014adam} as the optimizer and adaptive learning rate descent method as the learning rate scheduler. The initial learning rate is 1e-3, the patience is 10 epoch, the adjustment factor is 0.5, and the minimum learning rate is 1e-10. We trained the network for 100 epochs. In the test phase, the proposed network is a full-convolutional network, we input the source images without crop into the network to obtain the final fused image. The experiment is conducted on  the NVIDIA TITAN Xp and 128GB of CPU memory.
	
	\subsection{Fusion Evaluation} 
	Image quality evaluation is still a problem that has not been solved well. In this paper, we choose subjective evaluation and objective evaluation to evaluate the image quality.
	
	We compare the proposed method with some classic and latest fusion methods including cross bilateral filter fusion method(CBF)\cite{kumar2015image}, Laplacian Pyramid (LP)\cite{burt1983the}, Ratio of Low-pass Pyramid (RP)\cite{toet1989image}, Dual-Tree Complex Wavelet Transform (DTCWT)\cite{lewis2007pixel}, Curvelet Transform (CVT)\cite{nencini2007remote}, Multiresolution Singular Value Decomposition (MSVD)\cite{naidu2011image}, gradient transfer and total variation minimization(GTF) \cite{ma2016infrared}, CNN\cite{liu2017multi-focus}, FusionGan\cite{ma2019fusiongan}, Deepfuse\cite{prabhakar2017deepfuse} and DenseFuse\cite{li2018densefuse} respectively . 
	
	\begin{figure*}[!t]
		\includegraphics[width=\textwidth]{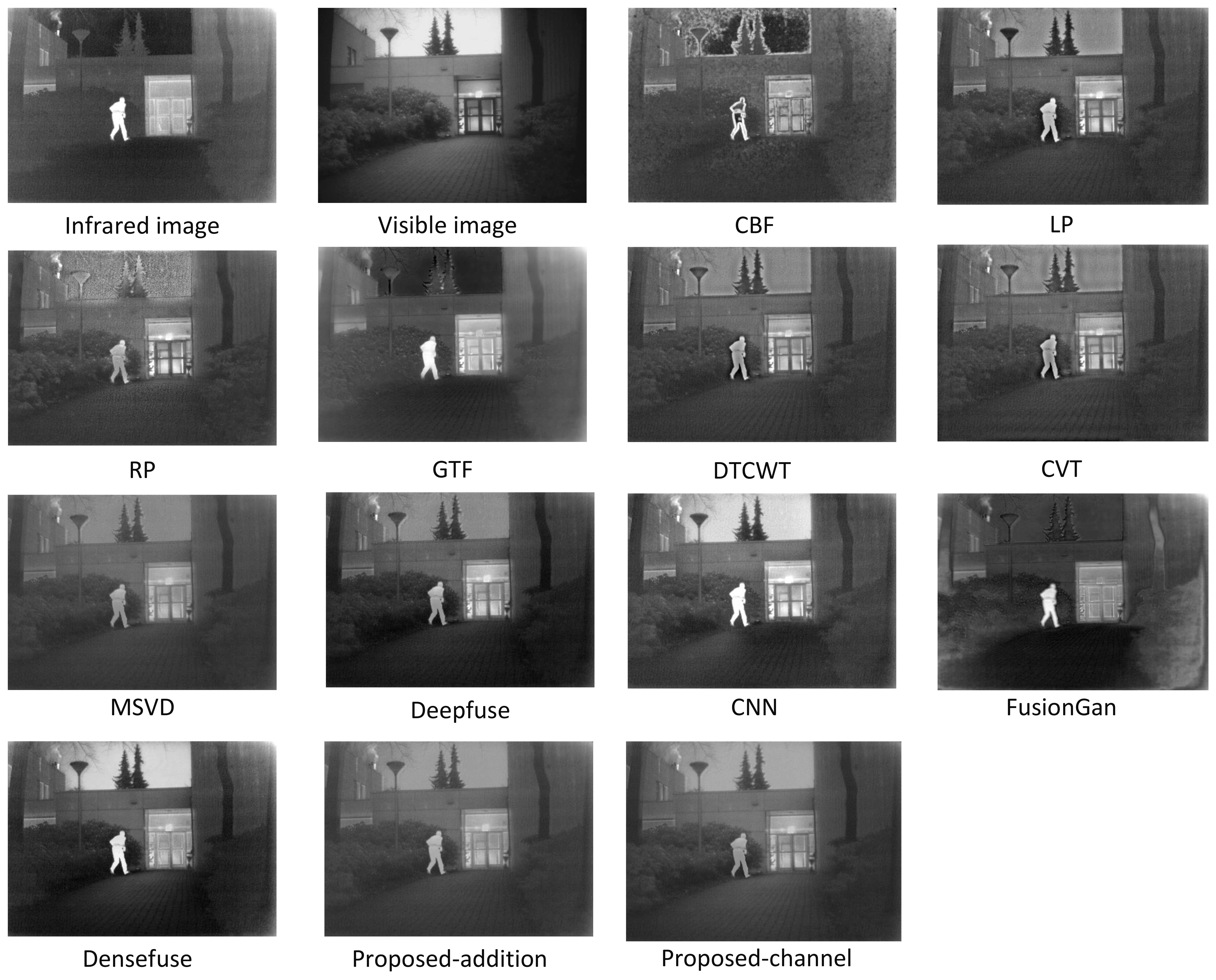}
		\caption{ Subjective evaluation of fused images of courtyard. Infrared image and visible light image are our input images. The other 11 images are the results of the comparison method and our proposed method.} \label{fig:images}
	\end{figure*}
	
	\subsubsection{Subjective Evaluation:}
	We follow human observation standards for images, such as brightness, contrast, color and other naturalness indicators. Generally speaking, it is the satisfaction degree of the image to people. In this paper, we list and compare the results of multiple algorithms.
	
	As shown in Fig. \ref{fig:images}, the channel strategy retains more detailed information such as the texture of bushes and global information such as the contrast of sky and buildings than the addition strategy. Compared with other methods, our proposed method also retains the details of visible light and infrared radiation information. At the same time, no noise and artifacts are introduced in the results. For example, noise points in the RP method, artifacts in CBF and FusionGan. In DenseFuse, the results are already quite good, but there are more white noises in the four corners.

	\subsubsection{Objective Evaluation:} 
	However, there is no single standard for subjective evaluation. People in different industries and different fields have different requirements for images. Therefore, we introduce five objective evaluation indicators, which are EI\cite{xydeas2000objective}, SF\cite{eskicioglu1995image}, EN\cite{roberts2008assessment}, SSIM\cite{wang2004image}, $N_{abf}$\cite{kumar2013multifocus} and MI\cite{qu2002information} respectively.
	
	EI represents the contrast intensity of adjacent pixels, SF represents the degree of mutation in the image, EN represents the amount of information contained in the image, SSIM represents the structural similarity between the two images, $N_{abf}$ represents the ratio of noise artificially added to the final image and MI represents the correlation of two images.
	\begin{table}[]
		\caption{Objective evaluation of classic and latest fusion algorithms. }\label{table:objective}
		\begin{center} 
			\resizebox{\linewidth}{20.2mm}{
			\begin{tabular}{|c|cccccc|}
				\toprule
				{\color[HTML]{000000} }              & {\color[HTML]{000000} EI}                        & {\color[HTML]{000000} SF}                        & {\color[HTML]{000000} EN}                       & {\color[HTML]{000000} SSIM}                     & {\color[HTML]{000000} Nabf}                     & {\color[HTML]{000000} MI}                        \\ \midrule
				{\color[HTML]{000000} CBF}           & {\color[HTML]{000000} \textit{\textbf{52.3045}}} & {\color[HTML]{000000} \textit{\textbf{13.0921}}} & {\color[HTML]{000000} 6.8275}                   & {\color[HTML]{000000} 0.6142}                   & {\color[HTML]{000000} 0.2669}                   & {\color[HTML]{000000} 13.6550}                   \\
				{\color[HTML]{000000} LP}            & {\color[HTML]{000000} 44.7055}                   & {\color[HTML]{000000} 11.5391}                   & {\color[HTML]{000000} 6.6322}                   & {\color[HTML]{000000} 0.7037}                   & {\color[HTML]{000000} 0.1328}                   & {\color[HTML]{000000} 13.2644}                   \\
				{\color[HTML]{000000} RP}            & {\color[HTML]{000000} 44.9054}                   & {\color[HTML]{000000} 12.7249}                   & {\color[HTML]{000000} 6.5397}                   & {\color[HTML]{000000} 0.6705}                   & {\color[HTML]{000000} 0.1915}                   & {\color[HTML]{000000} 13.0794}                   \\
				{\color[HTML]{000000} GTF}           & {\color[HTML]{000000} 35.0073}                   & {\color[HTML]{000000} 9.5022}                    & {\color[HTML]{000000} 6.5781}                   & {\color[HTML]{000000} 0.6798}                   & {\color[HTML]{000000} 0.0710}                   & {\color[HTML]{000000} 13.1562}                   \\
				{\color[HTML]{000000} DTCWT}         & {\color[HTML]{000000} 42.6159}                   & {\color[HTML]{000000} 11.1913}                   & {\color[HTML]{000000} 6.4799}                   & {\color[HTML]{000000} 0.6939}                   & {\color[HTML]{000000} 0.1428}                   & {\color[HTML]{000000} 12.9599}                   \\
				{\color[HTML]{000000} CVT}           & {\color[HTML]{000000} 43.1558}                   & {\color[HTML]{000000} 11.2006}                   & {\color[HTML]{000000} 6.5005}                   & {\color[HTML]{000000} 0.6907}                   & {\color[HTML]{000000} 0.1644}                   & {\color[HTML]{000000} 13.0011}                   \\
				{\color[HTML]{000000} MSVD}          & {\color[HTML]{000000} 27.9727}                   & {\color[HTML]{000000} 8.9758}                    & {\color[HTML]{000000} 6.2869}                   & {\color[HTML]{000000} \textit{\textbf{0.7219}}} & {\color[HTML]{000000} \textit{\textbf{0.0378}}} & {\color[HTML]{000000} 12.5738}                   \\
				{\color[HTML]{000000} Deepfuse}      & {\color[HTML]{000000} 33.8768}                   & {\color[HTML]{000000} 8.3500}                    & {\color[HTML]{000000} 6.6102}                   & {\color[HTML]{000000} 0.7135}                   & {\color[HTML]{000000} 0.0610}                   & {\color[HTML]{000000} 13.2205}                   \\
				{\color[HTML]{000000} CNN}           & {\color[HTML]{000000} 44.8334}                   & {\color[HTML]{000000} 11.6483}                   & {\color[HTML]{FF0000} \textbf{7.0629}}          & {\color[HTML]{000000} 0.6955}                   & {\color[HTML]{000000} 0.1286}                   & {\color[HTML]{FF0000} \textbf{14.1259}}          \\
				{\color[HTML]{000000} Fusion}        & {\color[HTML]{000000} 30.6847}                   & {\color[HTML]{000000} 7.5492}                    & {\color[HTML]{000000} 6.5299}                   & {\color[HTML]{000000} 0.6211}                   & {\color[HTML]{000000} 0.1344}                   & {\color[HTML]{000000} 13.0597}                   \\
				{\color[HTML]{000000} Densefuse}     & {\color[HTML]{000000} 36.4838}                   & {\color[HTML]{000000} 9.3238}                    & {\color[HTML]{000000} 6.8526}                   & {\color[HTML]{000000} 0.7108}                   & {\color[HTML]{000000} 0.0890}                   & {\color[HTML]{000000} 13.7053}                   \\
				{\color[HTML]{000000} ours-addition} & {\color[HTML]{000000} 25.3765}                   & {\color[HTML]{000000} 6.2758}                    & {\color[HTML]{000000} 6.2691}                   & {\color[HTML]{FF0000} \textbf{0.7593}}          & {\color[HTML]{FF0000} \textbf{0.0010}}          & {\color[HTML]{000000} 12.5382}                   \\
				{\color[HTML]{000000} ours-channel}  & {\color[HTML]{FF0000} \textbf{59.2286}}          & {\color[HTML]{FF0000} \textbf{21.9070}}          & {\color[HTML]{000000} \textit{\textbf{6.8793}}} & {\color[HTML]{000000} 0.5990}                   & {\color[HTML]{000000} 0.1958}                   & {\color[HTML]{000000} \textit{\textbf{13.7586}}} \\ \bottomrule
			\end{tabular}
		}
		\end{center}
	\end{table}
	
	It can be seen in Table. \ref{table:objective} that the result of the addition strategy obtains the best value in SSIM and $N_{abf}$, which means that this strategy retains a higher structural similarity and less noise. The result of the channel strategy obtains the best value in EI and SF, which means that the image contains more detailed edge information. The result of the channel strategy also has the second best value in EN and MI, which means that the result contains a lot of information from source image.
	
	Applying to addition or channel strategies, our algorithm has the best or second best value among the six indicators, which indicates that our proposed algorithm is effective for the fusion of infrared and visible light images.

	\section{CONCLUSION}
	In this paper, we propose a novel and effective deep learning network for infrared and visible light image fusion. The network we proposed has three parts, which are the encoder, fusion layer and decoder. After the two images are encoded, the extracted features are fused by the fusion layer, which is then input to the decoder to reconstruct the fused image. The encoder we designed has dual branch, one of which is the detail branch that uses dense connections to extract shallow layer information, and the other is the semantic branch that uses fast downsampling to extract global information. We design a new fusion strategy to fuse two sets of features according to the importance of the channel. Our proposed method has obtained the best value or the second best value on the six objective indicators. It shows that our proposed method has advantages for the fusion of infrared and visible light images. Although our research focuses on infrared and visible light image fusion, our proposed fusion strategy can be applied to a series of image fusion tasks including but not limited to medical image fusion and hyperspectral fusion. The proposed loss function can be used for a variety of image reconstruction or image generation tasks, and we hope the proposed network structure can be used for other image processing tasks.




	
	
	%
	
	\bibliographystyle{IEEEtran}
	\bibliography{mybibliography}
\end{document}